\documentclass{article} 
\usepackage{iclr2026_conference,times}

\usepackage{amsmath,amsfonts,bm}

\def\eqref#1{equation~\ref{#1}}

\def\1{\bm{1}}

\def\vc{{\bm{c}}}

\def\vx{{\bm{x}}}

\DeclareMathAlphabet{\mathsfit}{\encodingdefault}{\sfdefault}{m}{sl}
\SetMathAlphabet{\mathsfit}{bold}{\encodingdefault}{\sfdefault}{bx}{n}

\newcommand{\R}{\mathbb{R}}

\usepackage{hyperref}
\usepackage{url}
\usepackage{graphicx}

\usepackage{microtype}
\usepackage{graphicx}
\usepackage{subcaption}
\usepackage{booktabs}
\usepackage{amsmath}
\usepackage{amssymb}
\usepackage{mathtools}
\usepackage{amsthm}
\usepackage{hyperref}
\usepackage{multirow}
\usepackage{tikz}
\usepackage{algorithm}
\usepackage{algorithmic}
\usepackage{enumerate}
\usepackage{wrapfig}
\usepackage{tcolorbox}
\usepackage[frozencache]{minted}
\usepackage{adjustbox}
\usepackage{makecell}
\usepackage{xcolor}
\usepackage{enumitem}

\newcommand{\refig}[1]{Fig.~\ref{#1}}
\newcommand{\refapp}[1]{App.~\ref{#1}}

\newcommand{\refsec}[1]{Sec.~\ref{#1}}
\newcommand{\reftbl}[1]{Tab.~\ref{#1}}
\newcommand{\astar}{\ensuremath{A^*}}
\newcommand{\brackets}[1]{{\left<#1\right>}}
\newcommand{\braces}[1]{{\left\{#1\right\}}}

\newcommand{\mycolor}[2]{\textcolor{#1}{#2}}
\newcommand{\orange}[1]{\mycolor{orange}{#1}}
\newcommand{\blue}[1]{\mycolor{blue}{#1}}
\newcommand{\red}[1]{\mycolor{red}{#1}}
\newcommand{\green}[1]{\mycolor{green}{#1}}
\newcommand{\cyan}[1]{\mycolor{cyan}{#1}}
\newcommand{\magenta}[1]{\mycolor{magenta}{#1}}
\newcommand{\yellow}[1]{\mycolor{yellow}{#1}}

\title{``Don't Do That!'': Guiding Embodied Systems through Large Language Model-based Constraint Generation}

\iclrfinaltrue

\author{%
  Amin Seffo$^1$, Aladin Djuhera$^1$, Masataro Asai$^2$, Holger Boche$^1$ \\
  $^1$ Technical University Munich \texttt{\{amin.seffo, aladin.djuhera, boche\}@tum.de} \\
  $^2$ MIT-IBM Watson AI Lab \ \texttt{masataro.asai@ibm.com}
}

\begin{document}

\maketitle

\begin{abstract}
Recent advancements in large language models (LLMs) have spurred interest in robotic navigation
that incorporates complex spatial, mathematical, and conditional constraints from natural
language into the planning problem.
Such constraints can be informal yet highly complex, making it challenging 
to translate into a formal description that can be passed on to a planning 
algorithm.
In this paper, we propose STPR, a constraint generation framework
that uses LLMs to translate constraints (expressed as
instructions on ``what not to do'') into executable Python functions. 
STPR leverages the LLM's strong coding capabilities to shift the problem description 
from language into structured and interpretable code, thus circumventing complex reasoning 
and avoiding potential hallucinations.
We show that these LLM-generated functions accurately describe even complex mathematical 
constraints, and apply them to point cloud representations with traditional search algorithms. 
Experiments in a simulated Gazebo environment show that STPR ensures full compliance across
several constraints and scenarios, while having short runtimes.
We also verify that STPR can be used with smaller code LLMs, making it applicable 
to a wide range of compact models with low inference cost.
\end{abstract}

\section{Introduction}
\label{sec:introduction}

\begin{wrapfigure}{r}{0.48\textwidth} 
    \centering
    \includegraphics[width=1.0\linewidth]{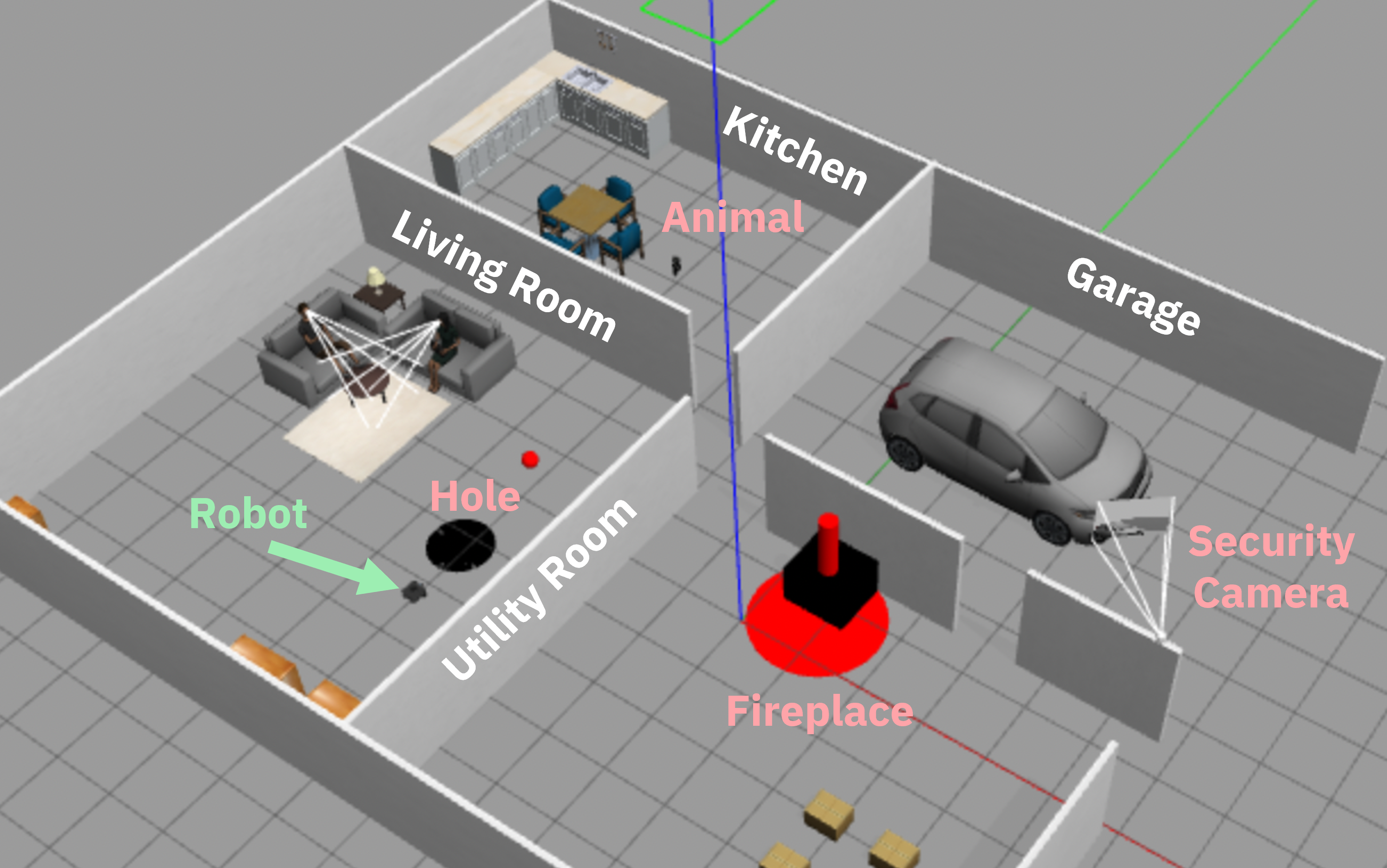}
    \caption{Gazebo environment with a garage, utility room, living room, and kitchen.
    A dangerous fireplace (red) must be avoided by the cleaning robot.}
    \label{fig:fireplace_example}
\end{wrapfigure}

Real-world navigation involves not only reaching a goal but also adhering to
constraints specified by human operators, which may be non-standardized, vague,
implicit, or informal, capturing \emph{semantic information} that is difficult
to extract from sensor data only.
For example, let us consider a cleaning robot without temperature sensors in \refig{fig:fireplace_example}.
The owner might instruct it to avoid getting close to a fireplace and may provide
additional details about its heat dissipation.
The challenge for the robot is to incorporate these contextual and potentially
complex spatial constraints into its path planning accordingly.

While recent advances in large language models (LLMs) have enabled robots to
interpret such natural language instructions \cite{huang2022language},
purely LLM-based planning,
where the language model directly returns a solution plan,
has several critical shortcomings:
First, LLMs can \textit{hallucinate}, generating seemingly plausible yet incorrect plans
that do not align with the robot's physical constraints or the true environment \cite{bubeck2023sparks}.
Second, non-reasoning models lack \textit{interpretability},
making it difficult to embed contextual constraints such as
site-specific hazards, social norms, or dynamically changing no-go zones.
Third, even if conditional constraints
(e.g., \textit{``never enter the kitchen if there is an animal''}) are directly encoded in an LLM prompt,
it often results in \textit{partial compliance}, i.e.,
LLMs may arbitrarily ignore or misinterpret constraints,
as they have no explicit mechanisms to enforce them reliably
\cite{liu2024deltadecomposedefficientlongterm}.
While advanced reasoning models (e.g., OpenAI o-family \cite{zhong2024evaluation}) may mitigate these weaknesses to some extent,
there is no theoretical guarantee that such issues never occur,
and yet these models incur significant computational cost and latency, both critical factors in real-time deployments.

\begin{figure*}[t]
    \centering
    \includegraphics[width=0.9\linewidth]{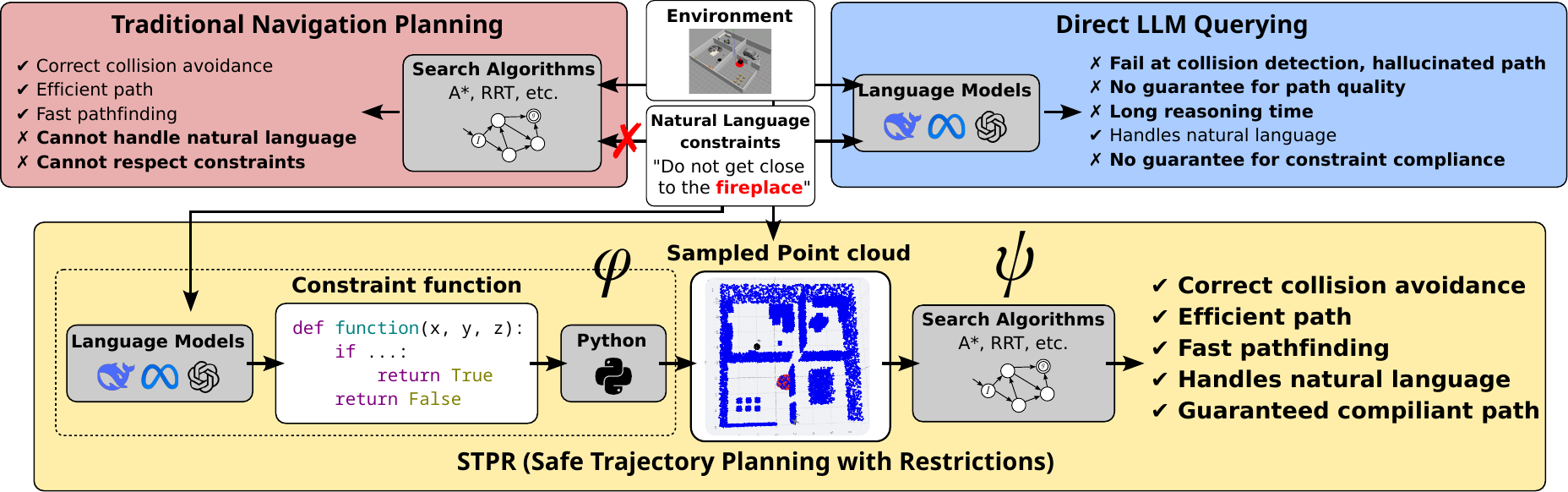}
    \caption{
    STPR Overview: LLM generates a Python function based on user constraints using a prompt template.
    The function is then integrated into a point sampling pipeline
    to generate a constrained representation of the environment.
    A classical algorithm uses this point cloud for path planning.}
    \label{fig:STPR_system}
    \vspace{-1em}
\end{figure*}
To address these limitations,
we propose \emph{Safe Trajectory Planning with Restrictions} (STPR, pronounced ``stopper'', \refig{fig:STPR_system}),
a practical and cost-effective neuro-symbolic navigation framework that
combines natural language comprehension with traditional pathfinding algorithms.
Instead of relying on LLMs for direct plan generation,
STPR employs LLMs to translate high-level natural language constraints (``what \emph{not} to do'')
into executable Python boolean functions.
These are then used to dynamically \emph{prune the robot's state space}
by generating point clouds that act as imaginary obstacles via a form of rejection sampling \cite{chib1995understanding}.
A traditional pathfinding algorithm
(e.g., \astar \cite{hart1968formal}, RRT* \cite{karaman2011sampling})
then operates within this refined state space, ensuring constraint-compliant navigation.

This approach bypasses potentially error prone complex mathematical or spatial
reasoning in text form, and instead leverages the LLM's strong pre-training on code to generate
precise, executable, and interpretable constraint functions.
It then relies on the rigorous theoretical guarantees of search algorithms,
such as optimality, soundness, and completeness.
This \emph{separation of concerns}, where LLMs handle constraint translation and
classical algorithms handle decision-making, allows STPR to circumvent common LLM
pitfalls while maintaining transparency in constraint enforcement.

We conduct a comprehensive evaluation of STPR in multiple \emph{Gazebo} \cite{gazebo} experiments with six different LLMs across four challenging scenarios, ranging from hazard avoidance to conditional safety rules.
Our empirical results show that STPR achieves \emph{full compliance}
and outperforms both simple vision-language model (VLM)-assisted
planners and established LLM-based planning frameworks, such as \emph{VoxPoser} \cite{voxposer},
in terms of success rate and plan quality.
Further, STPR does not require extensive hyperparameter tuning.
Thus, default inference settings suffice to produce robust constraint functions.
In addition, we verify that \emph{smaller code LLMs} are equally compatible with STPR and
generate reliable constraint functions without requiring larger, more sophisticated models.

\section{Problem Formulation and Preliminaries}
\label{sec:problem_formulation}

We define the robotic navigation problem
as a 4-tuple
$\Pi=\brackets{X, A, T, s_0, G}$,
where
$X\subseteq \R^3$ is a free space,
$A$ is a set of actions,
$T: X\times A\times X$ is a set of transitions,
$s_0\in X$ is the initial state,
and $G\subseteq X$ is the set of goal states.
A \emph{plan} is a sequence of actions $\pi=\brackets{a_1, a_2, \ldots a_N}$
such when executed (i.e., $\forall i; (s_i, a_i, s_{i+1})\in T$)
leads from $s_0$ to a goal state $s_N\in G$.
The corresponding sequence $\tau=\brackets{s_0,\ldots s_N}$ is then called a \emph{trajectory} or \emph{path}.
In this paper we use the latter terms interchangeably.
The quality of a plan is measured by its \emph{cost} and
computed as the sum of individual transition costs, i.e., $\sum_{t=0}^{N-1}\textit{cost}(s_t, a_t, s_{t+1})$.
For simplicity, we assume a Euclidean cost environment and consider a static, deterministic,
and discrete-time setting, though our model can be extended
to dynamic, stochastic, or continuous-time settings in the future.

We further model user-specified constraints as a mapping
$\phi: \mathcal{L} \rightarrow 2^X$,
where
$\mathcal{L}$ denotes the set of natural language instructions and
$2^X$ is a power set of $X$.
For any given instruction \(l \in \mathcal{L}\),
$\phi$ produces a \emph{forbidden region} $C\subseteq X \subseteq \R^3$
that must be avoided by the agent.
Herewith, we define the language-constrained robotic navigation problem as
$\Pi_l = \langle X', A, T', s_0, G' \rangle$,
where
$C = \phi(l)$,
$X' = X \setminus C$,
$T' = \{ (s, a, s') \in T \mid s, s' \in X' \}$,
and $G' = G \setminus C$.
We assume that the initial state \(s_0\) is not within the forbidden region \(C\).
A plan $\pi$ for $\Pi_l$ is called \emph{invalid} if a state
$s_t$ within its trajectory is part of $C$.
In the next section, we show how STPR leverages LLMs as a substitute for \(\phi\)
by translating human-specified instructions into Python constraint functions.

\section{Proposed STPR Method}
\label{sec:STPR_framework}

Our method is formalized as a meta-algorithm STPR($\phi,\psi$) that
takes an LLM-based constraint generator $\phi$ and a path finding algorithm $\psi$.
The LLM instantiates \(\phi\) by generating a Python function
\(f: X \rightarrow \{\mathrm{True},\mathrm{False}\}\),
which acts as an indicator function such that $C=\braces{\vx\in X\mid f(\vx)=\mathrm{True}}$.
\refig{fig:STPR_system} provides an overview of STPR where the
constraint functions augment the point-cloud representation of the environment.
We then employ $\psi$ to generate a constraint-compliant path for $\Pi_l$.
%
We will release all code, prompt templates, and implementation details as part of a public code repository (link to be provided upon acceptance).

\subsection{LLM-Based Constraint Code Prompting}
\label{sec:constraint_generation}

At the heart of STPR is a carefully engineered
\emph{prompt template} that elicits from the LLM a
self-contained Python function with a fixed signature (see \refig{fig:merged_prompt_output}).
Concretely, our template consists of four parts:

\begin{enumerate}[leftmargin=*]
    \item \textbf{System Instruction}: short directive (e.g., \emph{``You
    are a robot''}) that establishes the assistant's role and enforces
    zero-shot, single-turn behavior.

    \item \textbf{Environment Block}: textual representation of the environment. 
    Under simulation (e.g., in Gazebo), such a representation can be obtained by simply parsing the environment definition.
    In real-world deployments, object-recognition networks such as PointNet \cite{qi2017pointnet} or PointRCNN \cite{shi2019pointrcnn} can directly process raw point clouds to recover a similar symbolic scene representation.

    \item \textbf{Constraint Block}: human-readable description of the
    constraint (e.g., \emph{``Avoid the fireplace's heat dissipation
    zone''}) together with all relevant numeric parameters. such as object
    coordinates.

    \item \textbf{Python Signature}: function signature and doc-string
    ensuring the LLM never drifts into pseudocode/comments-only answers, thus producing a valid boolean test.
\end{enumerate}

By forcing the LLM to emit \emph{structured code} rather than free-form text,
we leverage its extensive pre-training on code to produce syntactically correct, idiomatic Python functions.
The rigid signature and minimal in-prompt examples thus leave little room for the model to stray,
effectively avoiding hallucinations.
Since every constraint can be provided directly by the user,
domain experts can audit or tweak it without complex retraining, making this approach highly generalizable.
STPR thus reduces the problem of \emph{``what not to do''} in natural language
to a transparent and executable Python predicate.


\subsection{Point-Cloud Sampling}
\label{sec:pointcloud_generation}

\begin{figure}[tb]
    \centering
    \footnotesize
    \begin{tcolorbox}[colback=gray!10, colframe=black, arc=1mm, boxrule=0.5pt, width=\linewidth, boxsep=-1.5mm]

        \textbf{STPR Prompt Template for Scenario 4: Fireplace Heat Avoidance}
        \vspace{0.5em}

        \orange{You are a robot placed in a house and need to get to a room with an active fireplace while maintaining safety distance to avoid heat.}


         The fireplace has the following parameters:
         x=0.5
         y=1.10
         z=0
         H\textunderscore0 = 1000 (heat intensity),
         alpha = 0.5  (dissipation factor),
         H\textunderscore safe = 50 (intensity threshold),
         d\textunderscore safe = 0.5  (distance threshold).

        \blue{Generate a safe geometric constraint by using
        a hemispherical heat radiation model and by following this
        python syntax without giving any usage example. Rename
        \texttt{<OBJ>} accordingly:}

        \begin{minted}[fontsize=\scriptsize]{python}
def is_in_constraints_OBJ(x, y, z):
  """
  Check if the input satisfies constraints.
  Parameters:
  - x, y, z (float): Point coordinates.
  Returns:
  - bool: True if point is forbidden, otherwise False
  """
        \end{minted}
    \end{tcolorbox}

    \vspace{2mm}
    {\Large $\Downarrow$} LLM
    \vspace{2mm}

    \begin{tcolorbox}[colback=gray!10, colframe=black, arc=1mm, boxrule=0.5pt, width=\linewidth, boxsep=-1.5mm]
        \begin{minted}[fontsize=\scriptsize]{python}
def is_in_constraints_fireplace(x, y, z):
    [...omitted...]
    # Calculate heat intensity using hemispherical model
    heat_intensity = H_0 / (4 * 3.14159 
                     * (distance ** 2)) * (1 - alpha)
    # Check if point is within fireplace boundaries
    within_fireplace = (
      abs(x - fireplace_x) <= fireplace_length / 2 and
      abs(y - fireplace_y) <= fireplace_width / 2 and
      abs(z - fireplace_z) <= fireplace_height / 2 )
    # Check if heat intensity exceeds safe threshold or 
    # distance is less than safe distance
    return within_fireplace or heat_intensity > H_safe 
                            or distance < d_safe
        \end{minted}
    \end{tcolorbox}

%
%
%
%
%

    \caption{\textbf{Top:} Prompt template for constraint code generation, including the System Instruction (\emph{``You are a robot ...''}) in \orange{orange}, the Environment Block in \textbf{black}, the Constraint Block with user-specific instructions (here: \emph{Scenario 4: Fireplace Heat Avoidance}) in \blue{blue}, and the Python function signature. \textbf{Bottom:} Corresponding constraint function generated by the LLM.}
    \label{fig:merged_prompt_output}
\end{figure}

STPR integrates LLM-generated constraints and other objects 
into a point cloud representation within a 3D environment, 
simulated using Gazebo.
This is done by a form of rejection sampling:
Let $o_i\in O$ be $i$-th static object (e.g., walls, furniture, etc.) in the environment.
To simplify the implementation,
each $o_i$ is over-approximated by a bounding box
$B_i=(\underline{x}_i, \overline{x}_i, \ldots, \underline{z}_i, \overline{z}_i)$
obtained by parsing an offline environment file,
i.e., a coordinate $\vx=(x,y,z)$ is regarded as colliding with $o_i$ when
$\underline{x}_i < x < \overline{x}_i$,
$\underline{y}_i < y < \overline{y}_i$,
and $\underline{z}_i < z < \overline{z}_i$.
Each $B_i$ thus implicitly represents a boolean function $f_i: X \rightarrow \{\mathrm{True},\mathrm{False}\}$
that returns \texttt{True} if a state $\vx\in X$ lies inside the bounding box.

Next, given a set of natural language instructions $\mathcal{L}$,
for each $i$-th instruction $l_i \in \mathcal{L}$,
we produce a boolean function $g_i$
of the same function signature using our prompt templates.
For each function $f_1, \ldots, f_{|O|}$ (represented by $B_1, \ldots, B_{|O|}$) and $g_1, \ldots, g_{|L|}$,
we generate a point cloud using rejection sampling:
For each $f_i$, we use $B_i$
to uniformly sample coordinates
$x\sim U(\underline{x}_i, \overline{x}_i), \ldots, z\sim U(\underline{z}_i, \overline{z}_i)$.
For each $g_i$,
we could uniformly sample points from the entire environment
and then reject them if $g_i(x,y,z)=\texttt{False}$.
However, such a naive rejection sampling can be inefficient due to the high rejection rate.
To address this issue,
we additionally query the LLM for a function that returns an over-approximating bounding box for the natural language constraint $l_i$,
sample some points from the bounding box, and then reject them using $g_i$.
In future work,
we might adapt a more sophisticated method such as \emph{adaptive rejection sampling} \cite{gilks1992adaptive}.

For each such function ($f_i$ or $g_i$),
we sample $K$ points in total (a hyperparameter in our setup)
and store them in a $k$d-tree structure \cite{bentley1975multidimensional} ($T_{f_i}$ or $T_{g_i}$)
for fast nearest neighbor queries during the subsequent planning step.
Note that the point cloud density may further affect the accuracy of when a constraint is violated.
We evaluate its impact in \refsec{sec:experiments_and_results}.

\subsection{Constrained Path Planning}
\label{sec:path_planning}

Given the initial state $s_0$ and the set of goal states $G$, represented by a goal condition
(e.g., within some Euclidean distance from the target coordinate),
STPR($\phi,\psi$) employs a generic path finding algorithm $\psi$.
In this paper, we apply \astar \cite{hart1968formal} on an 8-connected grid in the environment,
and RRT* \cite{karaman2011sampling} on a continuous space representation of the environment.
In general, $\psi$ can be replaced with more sophisticated and efficient algorithms
tailored toward continuous-space pathfinding under non-holonomic dynamics,
such as
Informed-RRT* \cite{gammell2014informed},
SST/SST* \cite{li2016asymptotically}, or others.
%


In our experiments, for either search algorithm,
each successor state $\vx$ is tested against the sampled points in the $k$d-trees.
Thus, given $\vx$, we query the nearest neighbor $\vc_{\text{nearest}}$, and,
if it is within the robot's maximum radius $R$,
i.e., $||\vx-\vc_{\text{nearest}}||_2 < R$,
then $\vx$ is pruned as a collision.
This means that collisions are only checked against the search nodes and not
along the edges because, in the more realistic scenario that involves a full physics simulation,
the trajectory generated by $\psi$ is used as navigation waypoints
for low-level local planners (e.g., Dynamic Window Approach \cite{fox1997dynamic} in ROS),
which find a collision-free path between them.

The search eventually comes to a stop when the expanded node is within $R$ of the goal,
it's priority queue becomes empty (in \astar), or when it hits the maximum iteration number (in RRT*).
Due to the completeness of \astar, the second case proves that no valid path exists on the 8-connected grid.
Further, since the search is guided by the Euclidean distance to the goal as the admissible heuristic,
the generated path is also guaranteed to be \emph{optimal} with regard to the path length on the 8-connected grid.
%
In RRT*, due to probabilistic completeness,
the chance of false negatives (reporting path nonexistence when it actually does exist) decays exponentially to the number of iterations.
Also, RRT* is asymptotically optimal, i.e.,
the cost of the solution \emph{almost surely} approaches the optimal cost
as the number of iterations tends to infinity.



We reiterate that in STPR \emph{constraint generation} is handled by the LLM,
while \emph{decision-making} remains within a well-understood planner.
This duality avoids pitfalls such as hallucinations or partial compliance by the LLM,
since $\psi$ only traverses states that are deemed valid.
As each constraint is an interpretable function, domain experts can modify
restrictions without retraining the LLM or altering the path planner.
Further, the use of point clouds as constraint representations is
a deliberate design choice, enabling STPR to augment existing visual
SLAM pipelines, for example, those that can be integrated within ROS's SLAM
Toolbox \cite{slamtoolbox} (see \refapp{app:ICRA_appendix}).

\def\dg{\textsuperscript{\textdagger}}

\section{Empirical Evaluations}
\label{sec:experiments_and_results}

We evaluate STPR across four challenging scenarios involving
spatial, conditional, and physical constraints.
We conduct all simulations in ROS within a Gazebo environment, including
a kitchen, living room, garage, and utility room (see \refig{fig:simulation_setup} in \refapp{app:ICRA_appendix}).
Each constraint is formulated in a \emph{``what not to do''}-style prompt written
by the human operator in addition to other relevant parameters.
%
The robot, based on a Turtlebot 3 Waffle,
has \emph{no sensors} that could aid in constraint handling,
as the purpose of our experiments is to show that
STPR allows for complementing the lack of sensor input
with user-specified natural language constraints.
For constraint generation, we use \emph{Llama-3.1-70B-Instruct} \cite{grattafiori2024llama3herdmodels} as our
primary model and explore additional models in the further analysis.
For point cloud sampling, we generate $K=1000$ points for each object $o_i$ and constraint $l_i$.
For $\psi$, we use STPR with both A* and RRT*,
referred to as STPR-A* and STPR-RRT*, respectively.

Our main interest is \emph{accuracy for path existence}
to evaluate two core properties of search algorithms:
\emph{Soundness}, i.e.,
the solution returned by an algorithm is guaranteed to be valid, and
\emph{completeness}, i.e.,
an algorithm should find a solution if there exists a solution,
or else report its nonexistence.
Out of $N=10$ runs, we measure the \emph{success ratio} by
counting a \emph{success}
whenever a method returns a valid solution for a solvable task or reports path nonexistence for an unsolvable task.
Otherwise we count it as a \emph{failure}.
In addition, we measure the total runtime until algorithm termination (including path nonexistence report),
as well as the length of the returned path as an indicator of solution quality.

\subsection{Simulation Scenarios}

\begin{figure*}[t]
    \centering
    \includegraphics[width=1\linewidth]{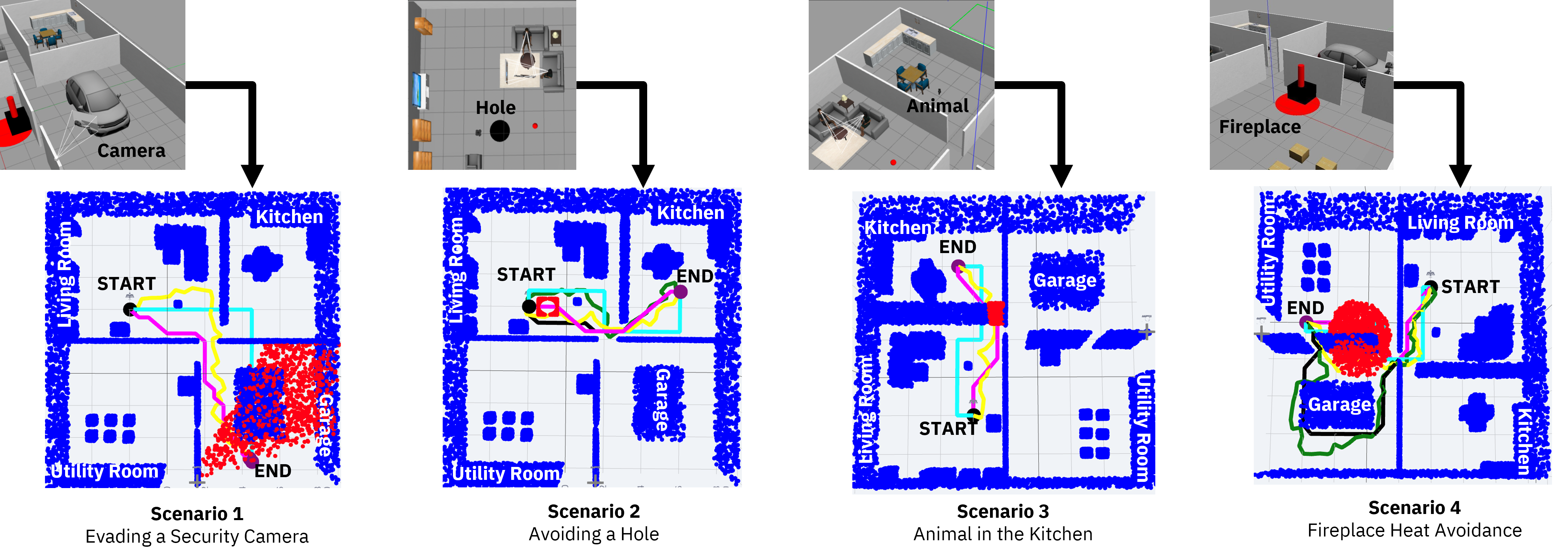}
    \caption{
    Planning results for STPR and baselines.
    \blue{\textbf{Blue}}: Point cloud for static objects.
    \red{\textbf{Red}}: Point cloud from constraint functions.
    \magenta{\textbf{Magenta}}: Path generated by vanilla \astar.
    \yellow{\textbf{Yellow}}: Path generated by vanilla RRT*.
    \textbf{Black}: Path generated by STPR-A*.
    \green{\textbf{Green}}: Path generated by STPR-RRT*.
    \cyan{\textbf{Cyan}}: Path generated by GPT-4o (using annotated image).
    Only STPR is compliant across all scenarios, refusing paths for \textbf{(S1)} and \textbf{(S3)}, and avoiding hazards for \textbf{(S2)} and \textbf{(S4)}.
    }
    \label{fig:scenarios}
    \vspace{-1em}
\end{figure*}

We consider four scenarios (see \refig{fig:scenarios}) that represent various types of constraints and complexities:

\begin{itemize}[leftmargin=*]

    \item \textbf{(S1) Evading a Security Camera.}
    The robot must avoid a security camera's field of view (FOV) defined by
    its projection parameters (position, yaw, and clip planes).
    This scenario challenges the system with visibility constraints that require
    complex mathematical modeling in 3D, relevant for privacy and stealth applications.

    \item \textbf{(S2) Avoiding a Hole.}
    The robot needs to avoid falling into an invisible/hidden pit trap.
    This tests the handling of non-obvious vertical hazards that even sophisticated sensors might miss
    due to the concealed nature
    (e.g., the trap could be covered by a carpet or overlooked by LiDAR sensors in occupancy grids),
    but it could be easily avoided with a simple verbal warning.

    \item \textbf{(S3) Animal in the Kitchen.}
    We instruct STPR that when an animal is present, the robot should avoid the kitchen.
    The environment has a small 3D model of a racoon in the kitchen area.
    We specifically label this encounter as a \emph{dangerous} event in our prompt template to see
    whether the LLM will correctly self-infer this conditional constraint from within the context.

    \item \textbf{(S4) Fireplace Heat Avoidance.}
    The robot must maintain a safe distance from a fireplace given its
    heat intensity, a safety threshold, and dissipation range.
    In this scenario, we test the LLM's ability to encode physics-based constraints beyond
    simple spatial boundaries.

\end{itemize}

To assess STPR's effectiveness, we deliberately plan paths between points
that would otherwise breach the constraints
and compare vanilla A*/RRT*, STPR, and naive VLM planning outcomes.
Here, \textbf{(S1)} and \textbf{(S3)} constitute unsolvable tasks
where a complete algorithm would report the nonexistence of valid paths
while a hallucinating algorithm would return a path that violates the constraints.
In \refsec{sec:voxposer}, we further compare STPR to \emph{VoxPoser}, an
LLM-based planning framework that generates affordance maps for constraint-aware navigation.

\subsection{Qualitative Results}



\paragraph{1) Vanilla \astar/RRT*}
As a sanity check, we first verify that
vanilla \astar and RRT* (without STPR)
indeed produce problematic paths in our scenarios.
The results are shown in magenta and yellow in \refig{fig:scenarios}.
In fact, all paths generated by both vanilla variants violate $\Pi_l$ and are thus deemed invalid.
For \textbf{(S1)}, the security camera is ignored and the agent enters the prohibited region.
For \textbf{(S2)}, the robot traverses over the hole as it lies on the optimal path
to the goal.
For \textbf{(S3)}, the kitchen is entered even though the animal is present, and
for \textbf{(S4)}, the trajectory gets too close to the fireplace.

\paragraph{2) STPR}
In contrast, for STPR-A*/-RRT*,
our experiments show \emph{full compliance} in all scenarios, where \astar/RRT* either
find a compliant path or correctly prove that no such path exists:

\begin{itemize}[leftmargin=*]
    \item For \textbf{(S1)},                    
    STPR derives a Python function that computes the distance and angular offset
    from the camera to each sampled point, approximating the FOV.
    %
    Specifically, the code checks if a point lies within the camera's near and
    far clip boundaries, and if its yaw-based angle remains inside the camera's
    horizontal FOV. 
    %
    %

    \item For \textbf{(S2)},                    
    a simple function is generated to check if a point falls into
    the hole with an added safety radius (visualized by the rectangular shape in red).
    %
    %

    \item For \textbf{(S3)},                    
    the constraint function constructs a forbidden region around the entrance,
    ensuring that \astar/RRT* cannot plan a path through the kitchen.
    Because this constraint is self-inferred based on the potentially dangerous encounter, STPR ensures
    context-aware planning, even though the environment is traversable (open door).
    This specific example shows effective reasoning where simply \emph{``closing''} the door
    is enough.

    \item For \textbf{(S4)},                    
    the constraint function implements a hemispherical heat radiation model,
    marking high-temperature zones as unsafe (see \refig{fig:merged_prompt_output}).
    This creates a safety radius, requiring \astar/RRT* to plan a detoured path which enters
    the utility room from another door.
    This example not only demonstrates an intelligent workaround but also that the LLM can produce a complex spatial constraint.
\end{itemize}


\paragraph{3) Naive VLM Planning}
Lastly, we evaluate a naive VLM approach that purely relies on
the model (GPT-4o/o3-mini-high) to generate an operational path from an annotated top-down
view of the environment (prompt and image will be provided in our code repository).
The resulting paths are shown in cyan in \refig{fig:scenarios}, alongside success ratios
and other metrics in \reftbl{tbl:comparison}.
In general, this approach suffers from hallucinations and partial compliance, where the model fails
to correctly recognize not only the constraints, but also static objects
or contexts within the environment.
Focusing on GPT-4o,
while there are some cases, such as \textbf{(S2)}, where the model succeeds at generating a valid path that
avoids the pit trap, paths for \textbf{(S1)} and \textbf{(S3)} are never valid, either violating the constraint,
going through objects (e.g., walls), or both.
This is particularly surprising as \textbf{(S3)} constitutes a simple conditional constraint that GPT-4o as a
reasoning model should be able to adhere to.
In addition, for \textbf{(S4)}, inconsistent outputs were observed, sometimes denying the existence of a valid path, yet
generating (an invalid) one.
The results for a high-performance reasoning model (GPT o3-mini-high)
are similar, while at significantly longer runtimes (see \reftbl{tbl:comparison}).
This shows that reasoning, even of advanced models, is brittle and
too unreliable for practical use in robotic path planning.

\begin{table*}[t]
    \centering
    \caption{Success ratio, runtime, and path quality of
    vanilla \astar/RRT*,
    STPR-A*/-RRT*, and
    naive VLMs (GPT 4o/o3-mini-high),
    averaged over $10$ runs.
    Valid best results in \textbf{bold},
    invalid results marked by \dg.
    }
    \label{tbl:comparison}
    \begin{adjustbox}{max width=\linewidth, scale=1}
    \begin{tabular}{l|cccccc|cccc|cccc}
        \toprule
        &
        \multicolumn{6}{c|}{\textbf{Success Ratio (\%)}} &
        \multicolumn{4}{c|}{\textbf{Total Runtime (s)}} &
        \multicolumn{4}{c}{\textbf{Path Quality (Length in meters)}} \\
         &
        \astar & RRT* & \makecell{STPR-\\A*} & \makecell{STPR-\\RRT*} & \makecell{GPT\\4o} & \makecell{GPT o3-\\mini-high} &
        \makecell{STPR-\\A*} & \makecell{STPR-\\RRT*} & \makecell{GPT\\4o} & \makecell{GPT o3-\\mini-high} &
        \makecell{STPR-\\A*} & \makecell{STPR-\\RRT*} & \makecell{GPT\\4o} & \makecell{GPT o3-\\mini-high} \\
        \midrule
        S1 & 0 & 0 & \textbf{100} & \textbf{100} & 0            & 0            & \textbf{14.17} &        {14.63}  & 34.10\dg & 238.50\dg & $\infty$       & $\infty$       & 20.0\dg & 20.4\dg \\
        S2 & 0 & 0 & \textbf{100} & \textbf{100} & \textbf{100} & \textbf{100} &        {12.51} & \textbf{11.97}  & 25.20    & 36.80     & \textbf{9.95}  &        {12.39} & 14.0    & 14.2    \\
        S3 & 0 & 0 & \textbf{100} & \textbf{100} & 0            & 10           & \textbf{12.91} &        {13.58}  & 24.30\dg & 105.20\dg & $\infty$       & $\infty$       & 12.0\dg & 12.3\dg \\
        S4 & 0 & 0 & \textbf{100} & \textbf{100} & 0            & 10           &        {18.01} & \textbf{17.31 } & 32.50\dg & 110.00\dg & \textbf{20.49} &        {25.80} & 14.7\dg & 15.1\dg \\
        \bottomrule
    \end{tabular}
    \end{adjustbox}
\end{table*}

\begin{table*}
  \centering
  \caption{Runtime breakdown across scenarios in seconds.
  \textbf{(Left)} Vanilla A*/RRT* runtimes. All paths are invalid (\dg).
  \\
  \textbf{(Right)} STPR-A*/-RRT* runtimes, including prompting, sampling, and planning.}
  \label{tab:scenario_results}
  \begin{adjustbox}{max width=\linewidth}
  \begin{tabular}{l | c c || c c c c c c}
    \toprule
    & \makecell{\textbf{Vanilla} \\ \textbf{A*}}
    & \makecell{\textbf{Vanilla} \\ \textbf{RRT*}}
    & \textbf{Prompting}
    & \textbf{Sampling}
    & \makecell{\textbf{Planning} \\ \textbf{(STPR-A*)}}
    & \makecell{\textbf{Planning} \\ \textbf{(STPR-RRT*)}}
    & \makecell{\textbf{Total} \\ \textbf{(STPR-A*)}}
    & \makecell{\textbf{Total} \\ \textbf{(STPR-RRT*)}} \\
    \midrule
    S1  & 0.60\dg & 0.01\dg & 12.84 & 0.13 & 1.20 & 1.66 & 14.17 & 14.63 \\
    S2  & 0.63\dg & 0.08\dg & 11.88 & 0.05 & 0.59 & 0.04 & 12.51 & 11.97 \\
    S3  & 0.63\dg & 0.08\dg & 11.77 & 0.03 & 1.11 & 1.78 & 12.91 & 13.58 \\
    S4  & 0.66\dg & 0.01\dg & 16.13 & 0.98 & 0.90 & 0.02 & 18.01 & 17.31 \\
    \bottomrule
  \end{tabular}
  \end{adjustbox}
\end{table*}

\subsection{Quantitative Results}

\paragraph{1) Runtime and Quality}
\reftbl{tbl:comparison} compares success ratios, total runtimes, and path lengths,
where results marked by \textdagger\ indicate that the metric belongs
to an invalid path due to collisions or violating constraints,
and where $\infty$ path lengths indicate that the method successfully
proves that no valid path exists.
While vanilla \astar and RRT* have shorter runtimes (see \reftbl{tab:scenario_results}), none of the paths adhere to the constraints,
as indicated by the 0\% success rate.
Similarly, for either GPT model, the VLM almost never succeeds in
generating a valid path (except for \textbf{(S2)}), with success rates between 0\% and 10\%.
In contrast, STPR returns paths with strong theoretical guarantees: optimality when using A*, and asymptotic optimality when using RRT*.
Further, paths returned by the VLM approach can be
either significantly longer than STPR's (e.g., 40\% for \textbf{(S2)} where the constraint is
respected) or
shorter (e.g., 28\% for \textbf{(S4)} where the constraint is neglected). However, the returned
waypoints are often too sparse and sometimes meters apart.
Such inconsistency shows another major weakness in end-to-end LLM-based planning.

\label{sec:sensitivity}


In \reftbl{tab:scenario_results},
we report detailed runtimes for the steps in STPR,
including prompting, point cloud sampling, and pathfinding.
In general, STPR maintains end-to-end latencies between 12 and 18 seconds, where
prompting takes some longer time for \textbf{(S4)}, in which
the LLM engages in detailed physical modeling.
For point cloud generation, \textbf{(S2)} and \textbf{(S3)} are faster due to their smaller forbidden regions,
while \textbf{(S1)} and \textbf{(S4)} require more time for the 3D spheres.
Depending on the goal state, planning varies, where \textbf{(S1)} and \textbf{(S3)}
take the longest for the search algorithms to determine that a valid path cannot be found.
Notably, for \textbf{(S2)} and \textbf{(S4)}, where valid paths do
exist, STPR's planning is significantly faster with RRT* due to its efficient
implementation.

\paragraph{2) Model Variations}
We ablated STPR's generative model across six different LLMs:
Llama-3.1-405B,
Llama-3.1-1B,
Granite-34B-Code \cite{mishra2024granitecodemodelsfamily},
GPT o1-pro, o3-mini-high, and 4o.
\reftbl{tab:inference_times} (left) shows the runtime for selected models that worked with STPR consistently.
The smallest Llama-3.1-1B (not shown in the table)
consistently fails at either logic or spatial reasoning, unable to generate constraints for any scenario.
A mid-tier code model such as Granite-34B-Code
produces sound functions, though requiring some minor prompt tuning to include missing Python imports.
As expected, advanced reasoning (GPT o1-pro/o3/4o) and top-tier (Llama-3.1-405B) models
achieve perfect success across all scenarios, however at the cost of higher prompting times.
For instance, o1 pro takes \emph{30 times longer} compared to our primary 70B Llama model, and
even more compared to the cost-effective Granite alternative.
This suggests that smaller code models can perform equally well.
Interestingly, we observe binary behavior such that
either the model works every time or never.
Thus, we omit success ratios from these results.
%
Note that the GPT entries in \reftbl{tbl:comparison} are for naive VLM-based planners, while \reftbl{tab:inference_times} compares runtimes for different LLM backends in STPR.

\paragraph{3) LLM Decoding Parameters}
We varied common LLM decoding parameters to examine the effect on STPR's performance.
In our exploratory testing,
STPR was robust against various top-$p$ values and temperature $\tau$ of nucleus sampling.
We observed prompting or sampling failures only when $(p,\tau)=(1,1)$,
which introduces excessive output randomness.
Further, default decoding settings (e.g., $(p,\tau) = (0.7, 0.2)$) in most APIs
sufficed to produce coherent constraint functions in our scenarios,
demonstrating that STPR does not require costly inference hyperparameter tuning.

\paragraph{4) Point cloud density}
This is a hyperparameter for STPR which
depends on the number of sampled points $K$.
A larger $K$ increases
the number of points in $k$d-trees and thereby
the average runtime $O(\log K)$ for each nearest neighbor query.
On the other hand, a smaller $K$ reduces the runtime, though
at the cost of potential failure to detect collisions.
To investigate this, we evaluated three values of $K=\braces{100,1000,10000}$
and measured the performance in \reftbl{tab:density-analysis} (right) for STPR-A*,
demonstrating that STPR's behavior is predictable and tuning $K$ is easy.

\begin{table*}[tb]
 \centering
 \caption{
 \textbf{(Left)} Prompting runtimes across LLMs
 (Llama-3.1-405/70B, Granite-34B-Code, GPT o1-pro, o3-mini-high, 4o).
 \\
 \textbf{(Right)} Accuracy vs. Runtime in STPR-A* (Llama-3.1-70B) for varying $K$.
 \textdagger\ indicates a run with an invalid path.
 }
 \label{tab:inference_times}
 \label{tab:density-analysis}
 \begin{adjustbox}{max width=\linewidth}
  \begin{tabular}{ccccccc|*{2}{|ccc}}
   \toprule
   & GPT
   & GPT o3-
   & GPT
   & Llama
   & Llama
   & Granite-
   & \multicolumn{3}{c|}{Success Ratio(\%)}
   & \multicolumn{3}{c}{Total Runtime(s)}
   \\
   & o1-pro
   & mini-high
   & 4o
   & 405B
   & 70B
   & 34B-Code
   & 100 & 1000 & 10000
   & 100 & 1000 & 10000
   \\\midrule
   S1 & 5m 40s & 1m 28s & 48.7s & 14.6s  & 12.8s & 13.1s &  0   & 100 & 100 & 12.9\dg & 14.1 & 22.6 \\
   S2 & 2m 32s & 1m 8s  & 35.0s & 27.7s  & 11.9s & 16.6s &  100 & 100 & 100 & 12.0    & 12.5 & 18.6 \\
   S3 & 6m 23s & 1m 35s & 39.5s & 23.7s  & 11.8s & 16.2s &  100 & 100 & 100 & 12.0    & 12.9 & 21.2 \\
   S4 & 7m 45s & 1m 55s & 52.2s & 21.7s  & 16.1s & 11.5s &  0   & 100 & 100 & 16.3\dg & 18.0 & 25.6 \\
   \bottomrule
  \end{tabular}
 \end{adjustbox}
\end{table*}
\section{Conclusion}

We proposed STPR, a neuro-symbolic robot navigation framework that leverages
LLMs to convert high-level natural language instructions into complex geometric constraints
expressed in Python functions.
STPR demonstrates that
it can quickly and reliably comply to diverse spatial, conditional, and physical constraints,
even when using smaller code LLMs.
We evaluated STPR using A* and RRT*,
demonstrating its compatibility with different search algorithms
that offer
(probabilistic) completeness, soundness, and (asymptotic) optimality.
We validated our approach through extensive simulations in ROS and ablation studies, covering four challenging scenarios, six different LLMs, and detailed runtime analyses.
Our work demonstrates that conversational LLMs can be reliably integrated into robot navigation,
while ensuring the properties of classical search algorithms. 
\subsubsection*{Acknowledgments}
This work was supported in part by the German Federal Ministry of Research, Technology and Space (BMFTR) within the research hub 6G-life (Grant 16KISK002), through the projects AISAC (Grant Number 16KIS2462) and 6G-Atlantic Bridges (in collaboration with MIT), by the Bavarian Ministry of Science and the Arts and the Saxon Ministry for Science, Culture, and Tourism through the project Next Generation AI Computing (gAIn), by the Bavarian Ministry of Economic Affairs, Regional Development and Energy through the project 6G Future Lab Bavaria, and in part by IBM Research.


\bibliography{iclr2026_conference}
\bibliographystyle{iclr2026_conference}

\newpage

\appendix
\section{Related Work}
\label{sec:related_work}

Similar to our work, some approaches attempt to improve safety, generate code, or use symbolic solvers, but the majority focus on high-level task planning or Task and Motion Planning (TAMP).
For example, VirtualHome \cite{huang2022language} and SayCan \cite{saycan2022arxiv} proposed using LLMs to translate high-level commands into actions, but validate feasibility via execution only, thus potentially performing unsafe actions.
Further, works such as Code-as-Policies \cite{Ajay2023CodeAsPolicies} and ProgPrompt \cite{singh2022progprompt} synthesize high-level scripts and programs, but cannot incorporate explicit low-level geometric constraints in continuous-space.
Symbolic specification methods such as DELTA \cite{liu2024deltadecomposedefficientlongterm}, LLM+P \cite{liu2023llmp}, LLMFP \cite{hao2025planning}, CaStL \cite{Guo2024CaStL}, and Thought-of-Search \cite{katz2024thought}, translate natural language into LTL, STL, PDDL, or SMT specifications, and then use classical planners or model checkers for sequencing.
However, logic-based representations consist of boolean variables (propositions) and operators over them (always, until, or, etc.), and thus LLMs can only generate the logic part. 
Further, the mapping of each proposition to a geometric object needs to be known which is not the case for our scenarios (e.g., camera's FOV does not exist as an object in Gazebo and is implicitly given via language). 
Thus, these approaches categorically cannot perform STPR’s low-level path planning nor synthesize complex geometric shapes.
In addition, extensions to TAMP, such as AutoTAMP, similarly generate STL specifications for temporal or chained reasoning, which is not the main focus of STPR.
In future work, it could be interesting to integrate STPR in TAMP systems as a single-goal subroutine. 
Further, in motion planning, PRoC3S \cite{Curtis2024PRoC3S} verifies LLM-generated motion plans using hand-coded state constraints, for example, through a Pybullet solver, thus not being able to generate geometric constraints.
Chance-constrained path planning (CCPP) \cite{blackmore2011chance} focuses on trajectory safety under stochastic dynamics, though, is orthogonal to our approach, as it assumes predefined obstacles.
However, integrating STPR's point clouds with CCPP is a promising future direction.
In this work, we further compared STPR to VoxPoser \cite{voxposer} which creates a volumetric costmap to shape a desired path.
In general, costmaps cannot express hard constraints as they lack soundness, thus they are strictly inferior and cannot report the nonexistence of paths unless there is a manually-tuned cost cap. 
In summary, STPR tackles low-level geometric navigation tasks under language-based constraints that require ad-hoc reasoning over complex geometric shapes. 
The significance of our work is showing that LLM code capabilities extend to low-level geometric navigation constraints, which has not been explored before.

\section{SLAM-based Simulation in ROS}
\label{app:ICRA_appendix}

\begin{figure}[htbp]
    \centering
    \includegraphics[width=0.6\linewidth]{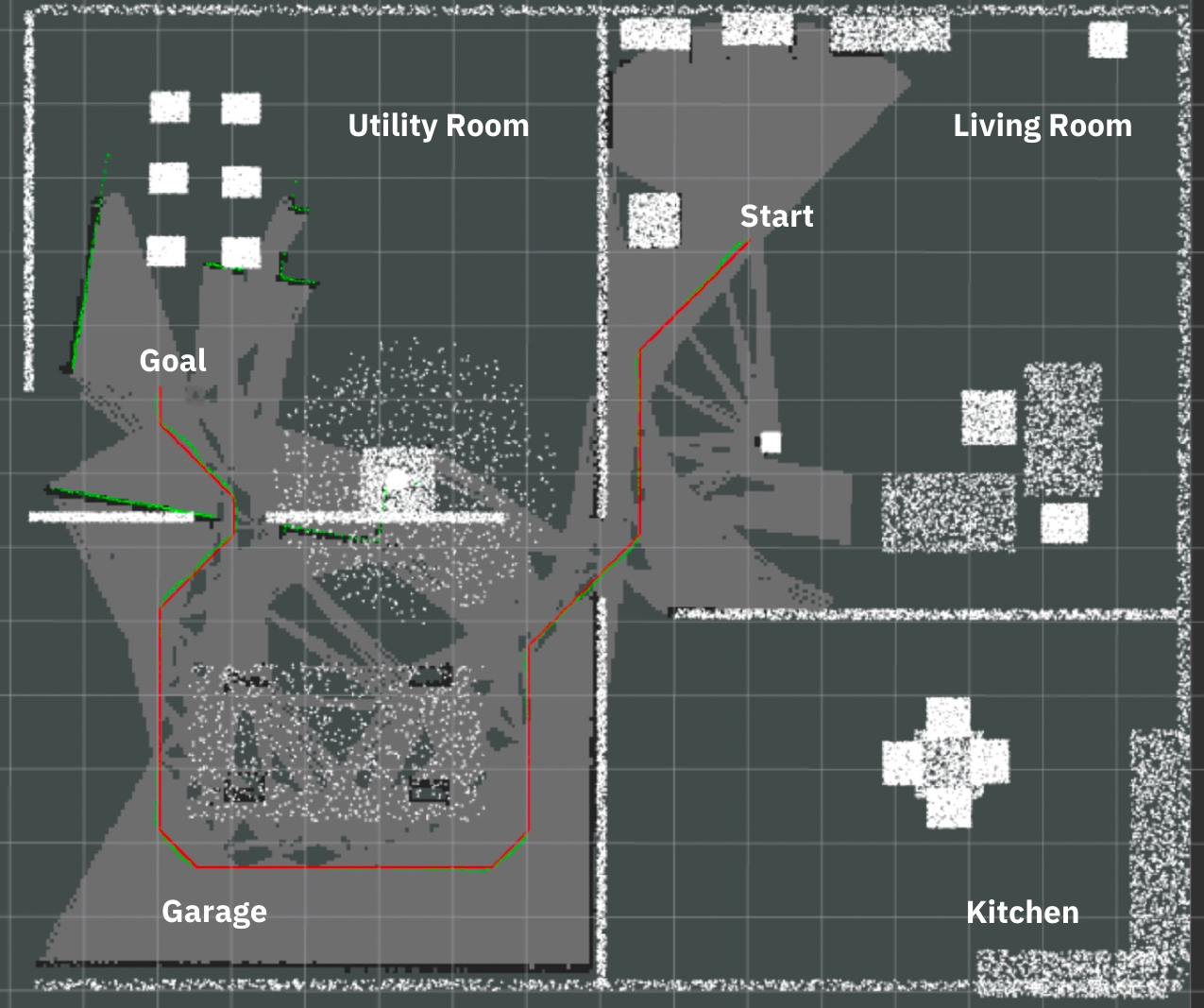}
    \caption{
        SLAM-generated occupancy grid map using LiDAR, visualized in RVIZ. Static structures such as walls and furniture are detected by the LiDAR, but potential hazards like the firecplace's heat zone cannot be captured by the sensor. The white points represent the point cloud generated by STPR. The \textcolor{red}{red} path represents the constraint-aware plan from STPR-A*, while the \textcolor{green}{green} line shows the executed ROS trajectory that avoids the unsafe region.
    }
    \label{fig:slam}
\end{figure}

While STPR is motivated by agents that lack the sensory capabilities to identify potential conflicts in their surroundings, we show in the SLAM map in \refig{fig:slam} that even with advanced LiDAR, certain hazards may remain undetected.
STPR bridges this gap by grounding high-level natural language instructions into constraint-aware plans, resulting in safe navigation using the constrained point clouds.
We perform simulations using off-the-shelf SLAM pipelines in ROS that internally handle SE(3), further showing a straightforward extension to SE(3).

\begin{figure}[t]
    \centering
    \includegraphics[width=0.95\linewidth]{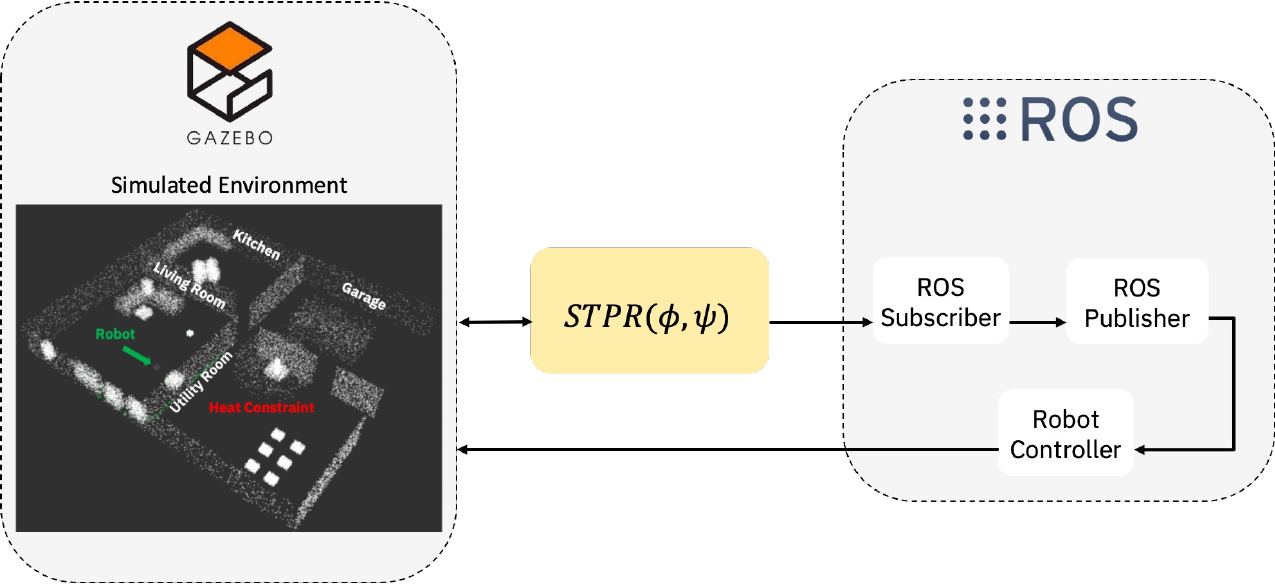}
    \caption{
        ROS setup for STPR: Natural language constraints are transformed into constraint functions that generate obstacle-aware point clouds for classical path planning. Planned trajectories are subscribed to by the robot, which executes the motion via published velocity commands in Gazebo.
    }
    \label{fig:simulation_setup}
    \vspace{-1em}
\end{figure}

\section{VoxPoser Replication}
\label{sec:voxposer}

\emph{VoxPoser} \cite{voxposer} is a method that uses LLMs to compose volumetric maps
that encode the cost for navigation as follows:
First, using an LLM,
it extracts task-relevant objects (e.g., \emph{``fireplace''}) from a list of static objects,
then writes code that invokes a VLM to detect object bounding boxes and object segmentation masks from a 2D image.
The masks are then projected onto 3D points obtained by multiple RGB-D cameras.
It then generates various volumetric cost maps using LLM-generated programs,
such as those for affordances and safety.
The maps are used as a policy of a trajectory planner.

We modified VoxPoser to our setup as follows.
First,
we allowed it a direct access to the ground-truth point clouds of the static objects,
as we did to STPR.
Instead of using LLM-generated programs,
we extract relevant object names using Llama-3.1-70B-Instruct \cite{grattafiori2024llama3herdmodels},
detect the corresponding objects in RGB images using Grounding DINO \cite{groundinDino} as a zero-shot object detection model,
use SAM 2 \cite{SAM2} to generate segmentation masks, project them 3D point clouds, and merge them with the static point cloud.
The robot has only one RGB-D camera instead of multiple,
which captures a single view of the scene at each time step.

Just as STPR, we generates a path using A*/RRT* guided by Euclidean distance heuristics, subject to the constrained point cloud.
We thus omit the generation of cost maps used by the greedy planner.
This is justified because cost maps are incapable of \emph{rejecting} a path,
even if allows the system to prefer a safer path.
For example, if all paths leading to the goal must go through certain high-cost voxels (i.e., unsafe region),
it has no mechanism to prevent a path through it and report path-nonexistence,
such as
a predefined or a predicted safety threshold on the values stored in each voxel.

\emph{The purpose of our experiment, therefore, is on refuting another critical property of VoxPoser, i.e., its \textbf{object-based paradigm.}}
Its entire pipeline is based on the belief that the environment consists of visual objects
and that they are sufficient for encoding complex natural language constraints.

\begin{figure}
 \centering
 \includegraphics[width=0.8\linewidth]{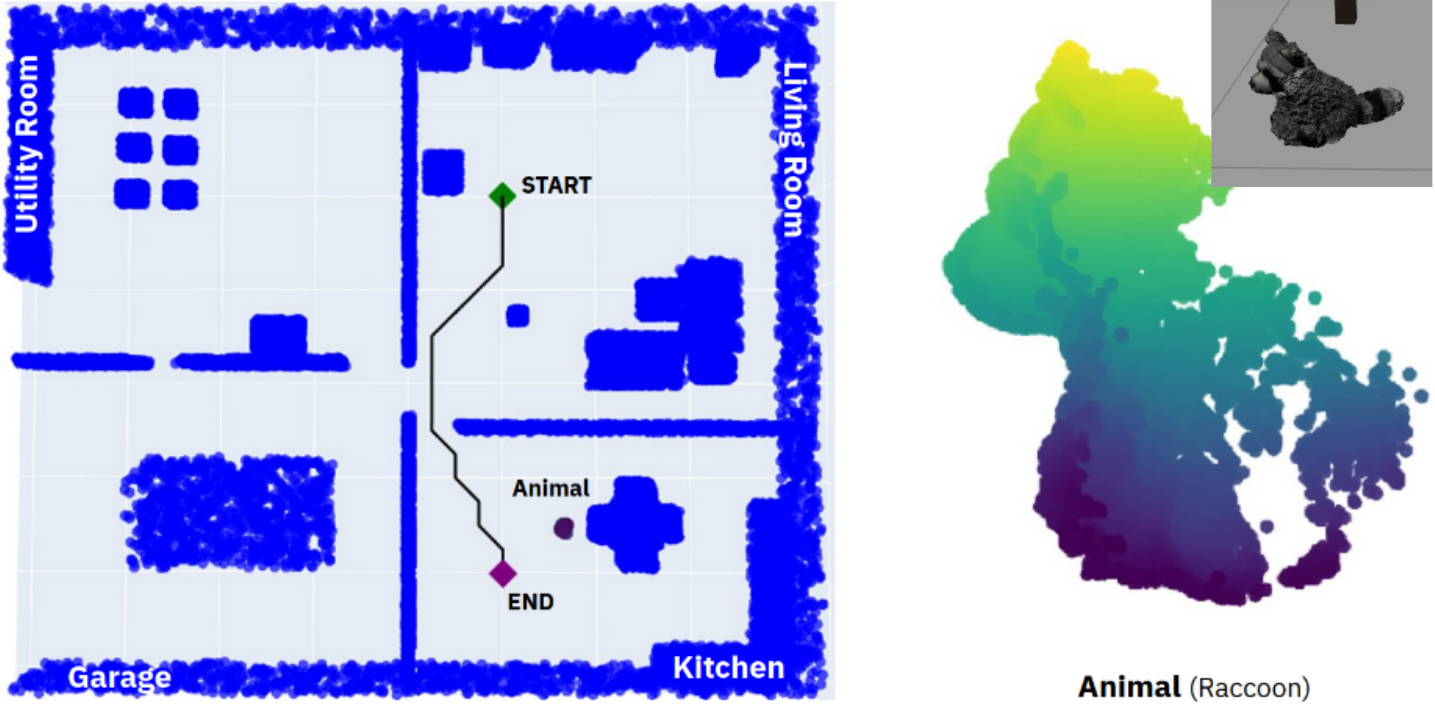}
 \caption{
 (Left) A path generated by VoxPoser in scenario \textbf{S3}, which should not have been generated according to the instruction.
 (Right) A constraint map encoding the safety,
 which was generated on the 2D camera image and projected on to the point cloud of a miniature 3D model of a racoon in the environment.
 Blue/Yellow represents $z$-axis value.
 }
 \label{fig:voxposer_animal}
 \vspace{-1em}
\end{figure}

%
%
The results are as follows:
\begin{itemize}[leftmargin=*]

    \item For \textbf{(S1)}, VoxPoser cannot be applied as it lacks support for abstract, invisible, or semantic constraints
          such as \emph{``avoid the security camera's FOV''} that cannot be treated as static objects observable by the VLM.

    \item For \textbf{(S2)}, the robot successfully deviates around the hole.
          This is the only scenario in which VoxPoser is able to plan a constraint-compliant path.
          While the path length is comparable to STPR, VoxPoser requires roughly twice the runtime due to the dense segmentation mask, which produces a significantly larger number of points.

    \item For \textbf{(S3)}, the racoon's shape (a dynamic object outside the static environment)
          is marked (\refig{fig:voxposer_animal}) as a set of unsafe voxels,
          but it still plans a path through the kitchen.

    \item For \textbf{(S4)}, it detects the fireplace, however, it cannot enforce the explicit heat radius and thus passes close by fireplace, eventually violating the constraint.

\end{itemize}

In comparison to STPR, which leverages callable constraint functions capable of modeling abstract, implicit, and semantic constraints, VoxPoser relies entirely on visual segmentation and therefore cannot handle such complex constraints.
This is particularly evident in \textbf{(S1)}, where STPR correctly avoids the cone-shaped camera frustum, which VoxPoser is unable to represent due to its dependence on segmented masks alone.
These results reinforce that STPR is more robust, modular, and interpretable, particularly for complex indoor environments requiring abstract reasoning with partial observability.

\end{document}